\documentclass{article}
\usepackage{ijcai26}

\usepackage{times}
\usepackage{soul}
\usepackage{url}
\usepackage[hidelinks]{hyperref}
\usepackage[utf8]{inputenc}
\usepackage[small]{caption}
\usepackage{graphicx}
\usepackage{subcaption}
\usepackage{amsmath}
\usepackage{amsthm}
\usepackage{amssymb}
\usepackage{booktabs}
\usepackage{algorithm}
\usepackage{algorithmic}
\usepackage[switch]{lineno}
\usepackage{multirow}
\usepackage{makecell}
\usepackage{adjustbox}
\usepackage{caption}

\title{TeNet: Text-to-Network for Compact Policy Synthesis}

\author{
Ariyan Bighashdel$^{1,2}$
\And
Kevin Sebastian Luck$^2$\\
\affiliations
$^1$Utrecht University\\
$^2$Vrije Universiteit Amsterdam\\
\emails
a.bighashdel@uu.nl,
k.s.luck@vu.nl
}

\begin{document}

\maketitle

\begin{abstract}
Robots that follow natural-language instructions often either plan at a high level using hand-designed interfaces or rely on large end-to-end models that are difficult to deploy for real-time control. We propose TeNet (Text-to-Network), a framework for instantiating compact, task-specific robot policies directly from natural language descriptions. TeNet conditions a hypernetwork on text embeddings produced by a pretrained large language model (LLM) to generate a fully executable policy, which then operates solely on low-dimensional state inputs at high control frequencies. By using the language only once at the policy instantiation time, TeNet inherits the general knowledge and paraphrasing robustness of pretrained LLMs while remaining lightweight and efficient at execution time. To improve generalization, we optionally ground language in behavior during training by aligning text embeddings with demonstrated actions, while requiring no demonstrations at inference time. Experiments on MuJoCo and Meta-World benchmarks show that TeNet produces policies that are orders of magnitude smaller than sequence-based baselines, while achieving strong performance in both multi-task and meta-learning settings and supporting high-frequency control. These results show that text-conditioned hypernetworks offer a practical way to build compact, language-driven controllers for ressource-constrained robot control tasks with real-time requirements.
\end{abstract}

\section{Introduction}

Recent advances in large language models (LLMs), such as GPT~\cite{brown2020language} and LLaMA~\cite{touvron2023llama}, have demonstrated that natural language can act as a powerful and flexible interface across a wide range of domains. In robotics, this has led to growing interest in language-conditioned control, where robots are guided by natural-language instructions, often alongside perceptual inputs. Prominent examples include vision–language–action (VLA) systems such as PaLM-E~\cite{driess2023palm}, SayCan~\cite{brohan2023can}, RT-2~\cite{zitkovich2023rt}, OpenVLA~\cite{kim2025openvla}, and OCTO~\cite{team2024octo}. These systems highlight how expressive language can be for specifying complex robotic behaviors. However, this expressiveness often comes at the cost of scale. \textbf{Many recent approaches to language-conditioned control rely on large end-to-end architectures, which can be computationally expensive and difficult to deploy in high-frequency control loops or on robots with limited onboard compute.}

At the other end of the spectrum are compact sequence models such as Decision Transformers (DT)~\cite{chen2021decision} and Prompt-DT~\cite{xu2022prompting}, which prioritize efficiency and ease of deployment in offline reinforcement learning. However, these models do not naturally incorporate language: instead, they rely on trajectory prompts or task-specific demonstrations to distinguish tasks, often requiring demonstrations even at test time and degrading as task diversity increases. This leaves a clear gap between expressive language-conditioned systems and compact policies that are efficient but are not language-enabled.

Several works attempt to bridge this gap by using language indirectly. Code-as-Policies~\cite{liang2023code} translates instructions into robot API calls, while Code-as-Rewards~\cite{venuto2024code} maps task descriptions into reward functions for reinforcement learning. Although effective in specific settings, these methods depend on predefined interfaces or accurate simulators, limiting their applicability to real-world robotics.

In this work, we ask a simpler question: \emph{can language itself be used as a direct conditioning signal for policy instantiation?} Rather than executing a large language model inside the control loop, we use it once—at policy instantiation—through a hypernetwork~\cite{ha2016hypernetworks}. We introduce \textbf{TeNet (Text-to-Network)}, a framework that conditions a hypernetwork on LLM-derived text embeddings to generate compact, task-specific policies. The resulting controller operates solely on low-dimensional state inputs, requires no demonstrations at inference time, and can run at high control frequencies on resource-constrained robots.

While direct text-to-policy instantiation is effective, we find that performance improves when language is grounded in behavior. By aligning language representations with expert trajectories during training, task descriptions capture not only linguistic intent but also behavioral semantics, leading to stronger generalization in multi-task and meta-learning settings. Importantly, grounding is used only during training: at inference, policies are instantiated from text alone.

This paper investigates a first approach into the possibility to utilze large robotic foundation models for resource-constrained robots via \emph{language-enabled hypernetworks for compact policy synthesis}. While this could be applied to VLAs in the future, our first steps focus on low-dimensional, trajectory-based domains (Mujoco and Meta-World), isolating the role of language in policy instantiation without addressing perception.

In summary, our contributions are:
\begin{itemize}
    \item \textbf{Text-to-Network Policy Generation.} We introduce TeNet, a framework that conditions a hypernetwork on LLM text embeddings to synthesize compact, task-specific robot policies suitable for real-time deployment.
    \item \textbf{Grounding Language in Behavior.} We show that aligning language with expert trajectories during training enriches linguistic representations with behavioral semantics and improves generalization in multi-task and meta-learning settings.
    \item \textbf{Empirical Insights into a New Paradigm.} We provide an extensive study on Mujoco and Meta-World benchmarks, highlighting both the promise and limitations of language-enabled hypernetworks and outlining paths toward future vision-grounded extensions.
\end{itemize}

\section{Related Work}

\textbf{Language-Conditioned Control in Robotics.}
Large language models (LLMs) have recently been integrated into robotic systems to enable natural language instruction following and high-level planning. Early approaches such as SayCan~\cite{brohan2023can} and PaLM-E~\cite{driess2023palm} leverage pretrained LLMs to map language into symbolic plans or action primitives executed by low-level controllers. These methods exploit LLMs’ world knowledge but typically operate at a goal or planning level rather than synthesizing executable control policies.

Other works connect language and control indirectly. Code-as-Policies~\cite{liang2023code} translates instructions into robot API calls, while Code-as-Rewards~\cite{venuto2024code} maps task descriptions into reward functions for reinforcement learning. SayTap~\cite{tang2023saytap} similarly maps language into structured locomotion patterns. While effective in constrained settings, these approaches rely on predefined interfaces or accurate simulators, limiting their generality. More recent vision-language-action systems, such as RT-2~\cite{zitkovich2023rt}, OpenVLA~\cite{kim2025openvla}, and OCTO~\cite{team2024octo}, integrate language and perception in large end-to-end models, but their computational demands hinder deployment in resource-constrained or high-frequency control settings.

Several works also explore grounding language in behavior through representation learning. For example, CLASP~\cite{rana2023contrastive} learns joint language–state–action embeddings via contrastive objectives, focusing on representation pretraining rather than policy synthesis. In contrast, our use of alignment is auxiliary: language grounding serves to stabilize text-conditioned policy generation rather than constituting the primary modeling objective.

\textbf{Compact Sequence Models for Policy Learning.}
A separate line of research explores compact sequence models as policies for reinforcement learning. The Decision Transformer (DT)~\cite{chen2021decision} formulates offline RL as conditional sequence modeling, generating actions autoregressively given states and return-to-go. While effective in single-task settings, DT lacks an explicit mechanism for task identification and therefore struggles in multi-task or meta-learning regimes.

Extensions such as Prompt-DT~\cite{xu2022prompting} and Meta-DT~\cite{wang2024meta} introduce task-conditioning via trajectory prompts, improving generalization at the cost of requiring demonstrations at inference time. Diffusion-based approaches, including MTDiff~\cite{he2023diffusion} and MetaDiffuser~\cite{ni2023metadiffuser}, similarly condition on trajectories or task contexts to generalize across tasks. Although these methods demonstrate strong performance, their reliance on prompt trajectories limits scalability when demonstrations are unavailable or expensive.

In parallel, visuomotor diffusion policies such as Diffusion Policy~\cite{chi2025diffusion} generate actions directly from images and have shown impressive real-world results. These approaches differ fundamentally from the low-dimensional, state-based settings we consider. We therefore focus on DT-based baselines to maintain architectural comparability and isolate the role of language-conditioned policy instantiation.

Overall, compact sequence models demonstrate that lightweight architectures can scale to multi-task RL, but their dependence on trajectory prompts and lack of direct language grounding constrain their applicability as instruction-following agents.

\textbf{Hypernetworks and Policy Generation.}
Hypernetworks~\cite{ha2016hypernetworks} generate the parameters of another network and have been widely explored for rapid specialization and meta-learning in reinforcement learning~\cite{beck2023hypernetworks}. Prior work conditions hypernetworks on a variety of signals, including structured task embeddings~\cite{rezaei2023hypernetworks,renhypogen}, demonstration trajectories~\cite{hegde2024latent,liang2024make}, behavior descriptors or archives~\cite{hegde2023generating}, robot morphology~\cite{xiong2024distilling}, or visual observations~\cite{gklezakos2022hyper}.

In parallel, language-conditioned hypernetworks have been studied in NLP to generate adapter or LoRA weights from task descriptions~\cite{ye2021learning,mahabadi2021parameter,lv-etal-2024-hyperlora,charakorn2025text}. These methods focus on adapting large language models rather than synthesizing control policies.

Across these domains, existing approaches either rely on structured task descriptors, demonstrations, morphology signals, or use language only to adapt large models. None directly combine LLM-based text encoders with hypernetworks to synthesize compact, task-specific robot control policies.

\textbf{Summary.}
Prior work has explored language-conditioned planning, compact sequence models, and hypernetwork-based policy generation. However, no existing approach directly instantiates executable robot policies from natural language via a shared hypernetwork. Our work fills this gap by using language as a conditioning signal for compact policy synthesis, grounded in behavior during training and executable without demonstrations at inference time.

\section{Problem Statement}

\textbf{Language-Augmented MDP (LA-MDP).}
We model a single task as a Language-Augmented MDP
\begin{equation}
\tilde{\mathcal{M}} = (\mathcal{S}, \mathcal{A}, P, R, \mu, H, \mathbb{L}),
\label{eq:la_mdp}
\end{equation}
which extends a standard MDP by including a language descriptor.  
The first six elements $(\mathcal{S}, \mathcal{A}, P, R, \mu, H)$ are the standard MDP components:  
$\mathcal{S}$ is the state space, $\mathcal{A}$ the action space, $P(s' \mid s,a)$ the transition dynamics,  
$R(s,a)$ the reward function, $\mu$ the initial state distribution, and $H$ the horizon.  
The additional component $\mathbb{L} \in \Delta(\mathcal{L})$ is a \emph{language descriptor}, i.e., a probability distribution over natural-language strings in the space $\mathcal{L}$.  
Each task is associated with its own descriptor distribution $\mathbb{L}$, which generates natural-language paraphrases 
(e.g., ``move forward'' vs.\ ``go straight'') of the same underlying dynamics $P$ and reward function $R$.  
Thus, the LA-MDP can be viewed as a standard MDP augmented with a generative source of equivalent task descriptions.  
A policy $\pi(a\mid s)$ induces a trajectory distribution in $\tilde{\mathcal{M}}$, and its performance is
\begin{equation}
J(\pi) = \mathbb{E}\!\left[\sum_{t=0}^{H-1} R(s_t,a_t)\right],
\label{eq:policy_perf}
\end{equation}
with the task-optimal policy $\pi^*=\arg\max_{\pi\in\Pi} J(\pi)$.

\textbf{Multi-task LA-MDP.}
We consider a distribution over tasks, where each task $\tau \in \mathcal{T}$ is an LA-MDP
\begin{equation}
\tilde{\mathcal{M}}_\tau = (\mathcal{S}_\tau, \mathcal{A}, P_\tau, R_\tau, \mu_\tau, H, \mathbb{L}_\tau).
\label{eq:multi_task_lamdp}
\end{equation}
Tasks may differ in $\mathcal{S}_\tau, P_\tau, R_\tau, \mu_\tau$ and $\mathbb{L}_\tau$, while sharing the action space $\mathcal{A}$. 
The multi-task objective is to learn a single policy that maximizes expected return across tasks: $\pi^* = \arg\max_{\pi\in\Pi} \;\mathbb{E}_{\tau \sim p(\mathcal{T})}\!\big[J_\tau(\pi)\big]$.

\textbf{Offline setting.}
No online interaction is permitted. 
The learner receives a static dataset collected from training tasks $\mathcal{T}_{\text{train}}$, each modeled as an LA-MDP
\begin{equation}
\mathcal{D}_{\text{train}} = \big\{\,(\mathcal{X}_\tau,\mathcal{D}_\tau)\;\big|\; \tau \in \mathcal{T}_{\text{train}} \big\},
\label{eq:offline_dataset}
\end{equation}
where $\mathcal{X}_\tau=\{\xi_\tau^{(k)}\}_{k=1}^{K}$ is a set of expert trajectories 
$\xi_\tau^{(k)}=(s_0,a_0,r_0,\ldots,s_H)$, 
and $\mathcal{D}_\tau=\{d_\tau^{(m)}\}_{m=1}^{M}$ are i.i.d. descriptions sampled from the language descriptor, $d_\tau^{(m)} \sim \mathbb{L}_\tau$.

\textbf{Multi-task learning.}
The learner is trained on demonstrations from a set of tasks $\mathcal{T}_{\text{train}}$.  
The objective is to learn a single model that approximates $\pi^*_\tau$ for all $\tau \in \mathcal{T}_{\text{train}}$, 
exploiting shared structure across tasks instead of training disjoint policies.  

\textbf{Meta-learning.}
The learner is trained on a collection of tasks $\mathcal{T}_{\text{train}}$ with the objective of generalizing to previously unseen tasks $\tau \in \mathcal{T}_{\text{test}}$.  
The challenge is to acquire transferable structure from $\mathcal{T}_{\text{train}}$ that enables rapid policy instantiation for new tasks without further environment interaction.  

\textbf{Few-shot adaptation (baselines).}
A common meta-RL strategy is to provide a small number of expert trajectories from the unseen task as adaptation data (few-shot setting). 
Prompt Decision Transformers (Prompt-DT) implement this by using short expert rollouts (\emph{prompt trajectories}) as test-time task identifiers.  

\textbf{Language-based instantiation (ours).}
In contrast Prompt-DT, we do not rely on prompt trajectories; instead we leverage natural-language descriptions sampled from $\mathbb{L}_\tau$ to instantiate policies for $\tau \in \mathcal{T}_{\text{test}}$, requiring the learner to ground language into behavior.

\section{Method}
\label{sec:method}
% \subsection{Overview}

\begin{figure*}[t]
\centering
\includegraphics[width=0.9\textwidth]{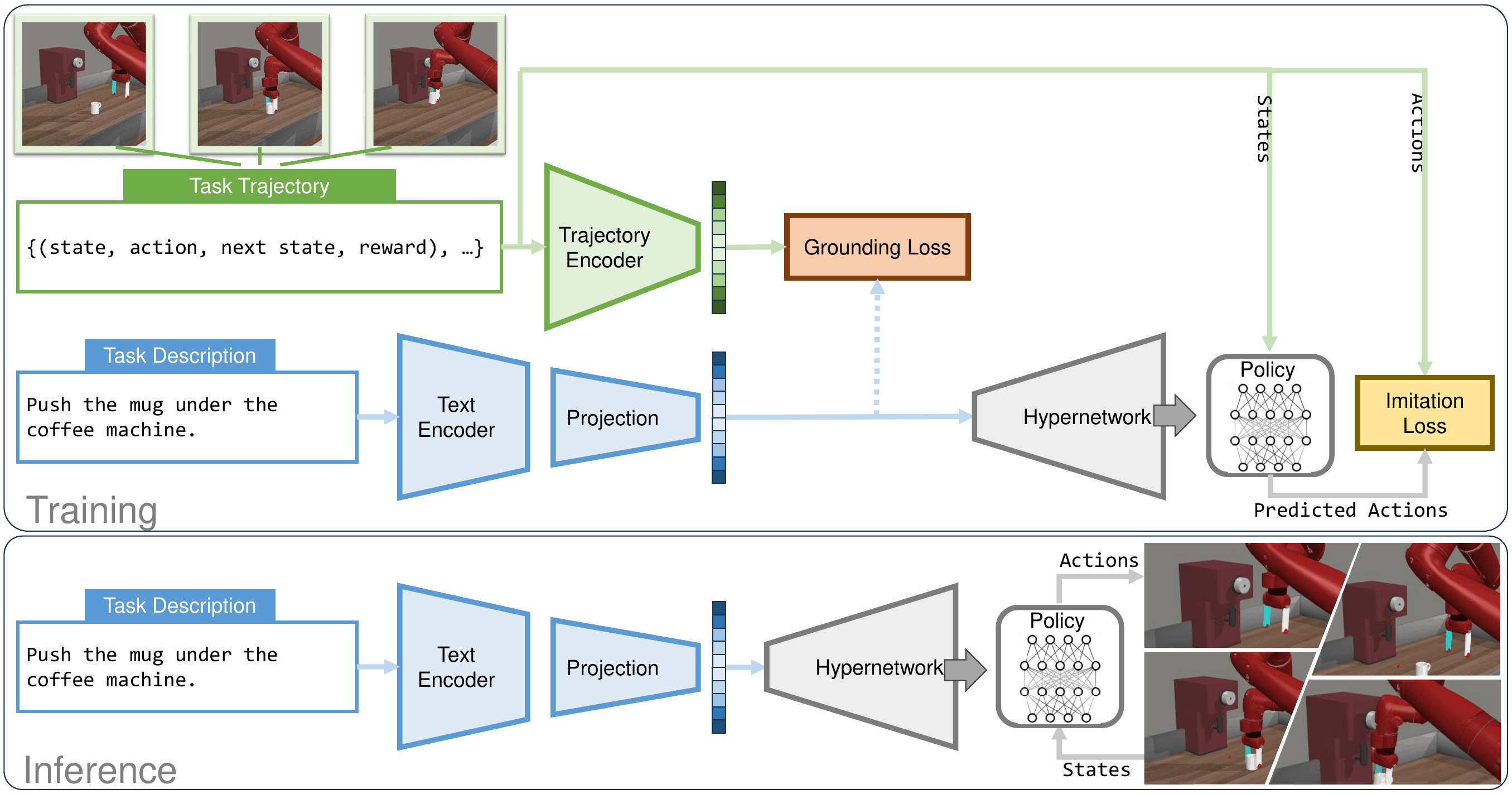}
\caption{Training (top) and inference (bottom) of the proposed framework. During training, trajectories and task descriptions are encoded, projected, and aligned through a language grounding module, with a hypernetwork generating task-specific policies optimized by imitation and grounding losses. At inference, only the task description conditions the hypernetwork to instantiate a policy that maps states to actions.}
\label{fig:framework}
\end{figure*}

Our framework, \textbf{TeNet (Text-to-Network)}, synthesizes compact, task-specific robot policies directly from natural language descriptions by conditioning a hypernetwork on language embeddings. At training time (Figure~\ref{fig:framework}, top), the model receives task descriptions and expert demonstrations. Task descriptions are first encoded into text embeddings. Expert demonstrations supervise the policy through an imitation loss. In the grounded variant, we additionally introduce a trajectory encoder, and align its embeddings with the text embeddings (i.e., language grounding), thereby enriching the language representation with behavioral semantics.
At inference time (Figure~\ref{fig:framework}, bottom), a new task description is passed through the text encoder, projected to the appropriate embedding space, and fed into the hypernetwork to generate a policy that can be executed without further demonstrations.  

We present two variants of our approach: \textbf{Direct TeNet}, which conditions the hypernetwork solely on text embeddings, and \textbf{Grounded TeNet}, which aligns text embeddings with trajectory embeddings during training to capture behavioral semantics and improve generalization.

\subsection{Direct TeNet}
\label{sec:direct-tenet}
In the Direct TeNet variant, policies are instantiated directly from task descriptions without trajectory grounding. Given a description $d \in \mathcal{L}$, the text encoder $f_{\text{text}}$ produces an embedding $z_d = f_{\text{text}}(d) \in \mathbb{R}^{d_z}$. A projection network $g$ maps $z_d$ into the conditioning space of the hypernetwork: $\tilde{z}_d = g(z_d)$. The hypernetwork $h$ then generates the parameters $\theta_\pi$ of a task-specific policy network $\pi_{\theta_\pi}$
\begin{equation}
\theta_\pi = h(\tilde{z}_d), \qquad \pi_{\theta_\pi}(a \mid s).
\label{eq:hypernetwork}
\end{equation}

Training relies on expert demonstrations $\xi_\tau = \{(s_t,a_t)\}_{t=0}^{H}$ from task $\tau$. The policy is supervised by behavior cloning (imitation learning)
\begin{equation}
\mathcal{L}_{\text{BC}} = - \mathbb{E}_{(s,a) \sim \xi_\tau} \big[ \log \pi_{\theta_\pi}(a \mid s) \big].
\label{eq:bc-loss}
\end{equation}

Thus, Direct TeNet provides a simple mechanism for mapping language directly into executable policies through the hypernetwork. 

\subsection{Grounded TeNet}
\label{sec:grounded-tenet}
Direct TeNet instantiates policies solely from projected text embeddings (Section~\ref{sec:direct-tenet}). 
To better capture behavioral semantics, Grounded TeNet augments training with additional grounding objectives that align text and trajectory embeddings. We emphasize that grounding is not the primary conceptual contribution of TeNet: it is an auxiliary mechanism that stabilizes and enriches the text embeddings, while the core novelty lies in generating executable policy parameters directly from natural language.

Given an expert trajectory $\xi = \{(s_t,a_t,r_t,s_{t+1})\}_{t=0}^{H}$, the trajectory encoder 
$f_{\text{traj}}$ produces an embedding $z_\xi = f_{\text{traj}}(\xi)$. Both $z_\xi$ and the projected text embedding $\tilde{z}_d$ are mapped into a shared space, 
and a grounding loss $\mathcal{L}_{\text{ground}}$ is applied. We explore two variants:

\paragraph{Direct alignment (MSE).}
A simple strategy is to directly minimize the squared distance between projected text and trajectory embeddings $\mathcal{L}_{\text{align}}
= \mathbb{E}_{(d,\xi)}
\big[ \| \tilde{z}_d - z_\xi \|_2^2 \big]$.
This objective enforces absolute closeness of paired embeddings in the shared space.  

\paragraph{Contrastive alignment.}
Let $\mathrm{sim}(\cdot,\cdot)$ denote cosine similarity and $\beta>0$ a temperature parameter. 
For each update, we consider a finite candidate set of trajectory embeddings 
$\mathcal{C}_\xi$ and a finite candidate set of text embeddings $\mathcal{C}_d$ 
that provide negatives for the contrastive normalization.

\emph{(i) Text–trajectory contrastive (symmetric).}
For paired $(\tilde{z}_d, z_\xi)$, we align text to trajectory and trajectory to text with a symmetric InfoNCE
\begin{equation}
\begin{aligned}
\mathcal{L}_{\text{text-traj}}
= \tfrac{1}{2}\,\mathbb{E}_{(d,\xi)} \Bigg[
&-\log \frac{\exp\big(\mathrm{sim}(\tilde{z}_d, z_\xi)/\beta\big)}
{\sum_{\xi' \in \mathcal{C}_\xi} \exp\big(\mathrm{sim}(\tilde{z}_d, z_{\xi'})/\beta\big)} \\
&-\log \frac{\exp\big(\mathrm{sim}(\tilde{z}_d, z_\xi)/\beta\big)}
{\sum_{d' \in \mathcal{C}_d} \exp\big(\mathrm{sim}(\tilde{z}_{d'}, z_\xi)/\beta\big)}
\Bigg].
\end{aligned}
\label{eq:text-traj-symmetric}
\end{equation}

\emph{(ii) Text–text contrastive.}
Task descriptions can be structurally similar (e.g., differing only in goal parameters), 
which may collapse text embeddings. To encourage description-level discrimination, we add
\begin{equation}
\mathcal{L}_{\text{text-text}}
= \mathbb{E}_{d}
\Bigg[
-\log \frac{\exp\big(\mathrm{sim}(\tilde{z}_d, \tilde{z}_d)/\beta\big)}
{\sum_{d' \in \mathcal{C}_d} \exp\big(\mathrm{sim}(\tilde{z}_d, \tilde{z}_{d'})/\beta\big)}
\Bigg].
\label{eq:text-text}
\end{equation}

The final contrastive objective is $\mathcal{L}_{\text{contrastive}}
= \mathcal{L}_{\text{text-traj}} + \mathcal{L}_{\text{text-text}}$.

\paragraph{Summary.}
The total training loss combines imitation learning with grounding: $\mathcal{L} = \mathcal{L}_{\text{BC}} + \lambda_{\text{g}} \, \mathcal{L}_{\text{ground}}$, where $\mathcal{L}_{\text{ground}}$ may include $\mathcal{L}_{\text{align}}$ or 
$\mathcal{L}_{\text{contrastive}}$, 
and $\lambda_{\text{g}}$ balances their contribution. At inference time, no trajectories are required -- the policy is instantiated from text alone. 
Grounding is used only during training to shape the representation.

\section{Experiments}
\label{sec:experiments}
We conduct an extensive empirical study to evaluate TeNet and to provide insights into the design 
and behavior of language-enabled hypernetworks. Our experiments are performed on Mujoco control 
benchmarks (HalfCheetah-Vel, HalfCheetah-Dir, Ant-Dir)~\cite{todorov2012mujoco} and Meta-World manipulation benchmarks 
(ML1 Pick-Place, MT10, MT50)~\cite{pmlr-v100-yu20a}, covering both multi-task and meta-learning settings. 

Beyond reporting standard performance, our goal is to systematically answer a series of questions 
about when and why TeNet is effective, how grounding influences policy quality, and how design choices affect performance. 
This section is therefore organized around these questions, with results interleaved with analysis.
\subsection{Experimental Setup}

\paragraph{Benchmarks.}
We evaluate on \emph{Mujoco} locomotion (HalfCheetah-Dir, HalfCheetah-Vel, Ant-Dir) and \emph{Meta-World} manipulation (ML1 Pick-Place, MT10, MT50), spanning multi-task and meta-learning regimes. Full task definitions, state/action spaces, and splits are provided in the supplementary materials.

\paragraph{Models.}
We compare \textbf{DT}~\cite{chen2021decision}, \textbf{Prompt-DT}~\cite{xu2022prompting}, and three TeNet variants: \textbf{TeNet} (direct, no grounding), \textbf{TeNet-MSE} (MSE grounding), and \textbf{TeNet-Contrast} (contrastive grounding). Implementation details, Prompt-DT size variants, and the Prompt-DT+Hypernetwork modification are provided in the supplementary materials.

\paragraph{Metrics \& protocol.}
We report \emph{episodic return} on Mujoco and \emph{success rate} on Meta-World, plus \emph{controller size} and \emph{control frequency} for deployability. Results are averaged over 3 seeds; each task is evaluated with 50 rollouts.

\paragraph{Defaults.}
Unless stated otherwise: the text encoder is \emph{Llama-3 8B} (frozen), the trajectory encoder is \emph{Prompt-DT} (used only for grounded variants), and TeNet uses a small MLP hypernetwork to instantiate a $\sim$40K-parameter policy. Training is strictly offline. 

\subsection{Results}
\label{sec:main-results}

\begin{figure*}[h]
\centering
\includegraphics[width=0.95\textwidth]{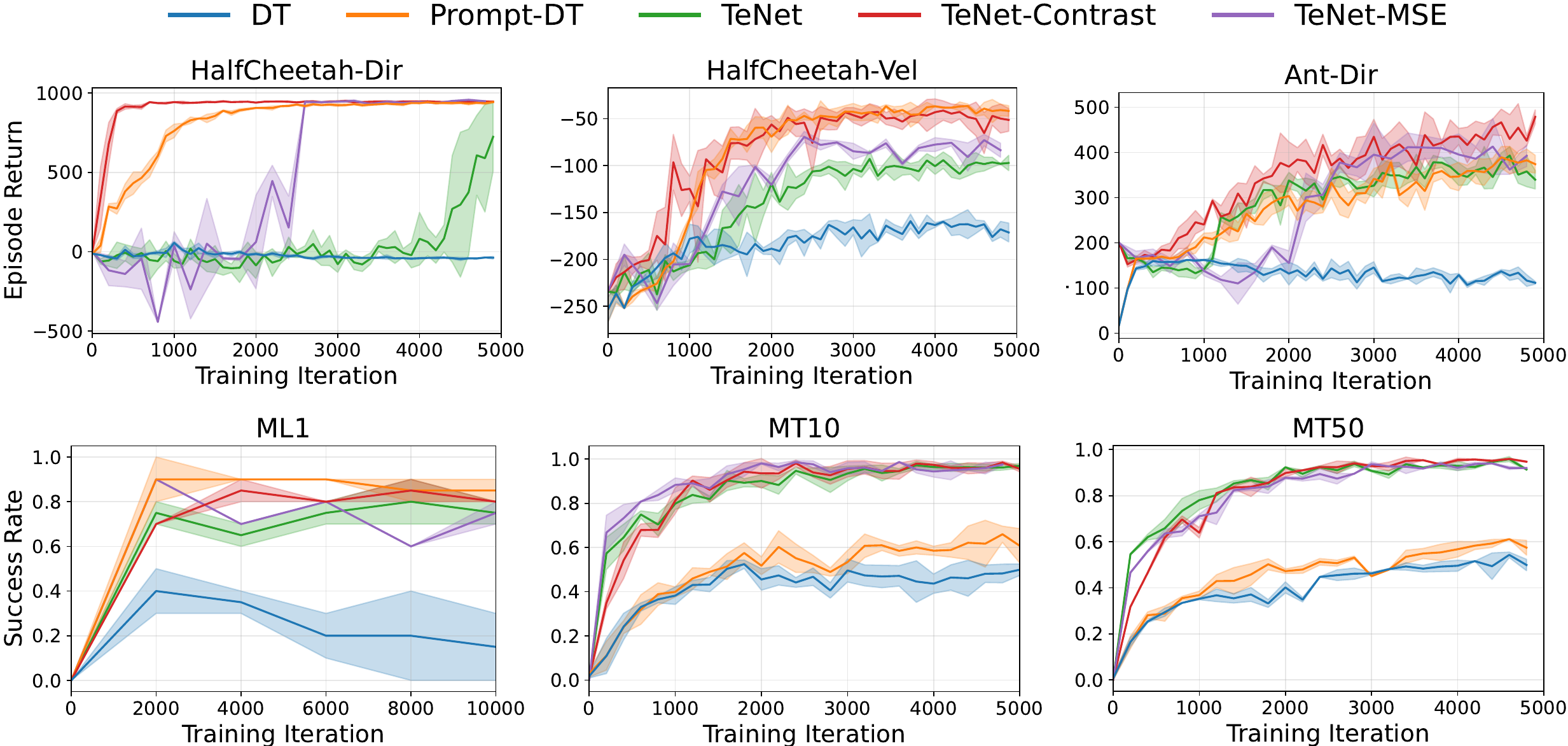}
\caption{Performance across Mujoco (HalfCheetah-Dir, HalfCheetah-Vel, Ant-Dir) 
and Meta-World (ML1 Pick-Place, MT10, MT50). 
Each subplot reports mean and standard deviation over three seeds. 
A shared legend is shown at the top.} 
\label{fig:main-results}
\end{figure*}

Figure~\ref{fig:main-results} summarizes performance across all six benchmarks, 
with a shared legend shown on top. 

Several general trends are clear. 
First, \textbf{DT} is consistently the weakest model across all domains, 
confirming that a compact sequence model without explicit task signals 
is not suitable for multi-task or meta-learning. 
Both \textbf{Prompt-DT} and \textbf{TeNet} address this limitation by 
providing task signals, but they do so in fundamentally different ways: 
Prompt-DT relies on short expert rollouts (prompt trajectories) as identifiers, while TeNet derives task signals directly from natural language descriptions. This text-based conditioning avoids the need for demonstrations at test time, making TeNet more scalable and practical \emph{within our state-based multi-task benchmarks}, as it removes the requirement for task-specific trajectory prompts.

Second, when comparing \textbf{TeNet} variants (more specifically \textbf{TeNet-Contrast}) against \textbf{Prompt-DT}, we observe consistent advantages. 
TeNet-Contrast outperforms Prompt-DT in HalfCheetah-Dir and Ant-Dir, 
matches it in HalfCheetah-Vel, and is slightly worse in ML1 Pick-Place 
(which we analyze further in Section~\ref{sec:task_scaling}). 
Most strikingly, in MT10 and MT50 TeNet-Contrast \emph{hugely outperforms} Prompt-DT. 
This large gap prompted us to investigate why Prompt-DT struggles so severely in 
multi-task benchmarks and to identify which design choices in TeNet 
are responsible for its robust performance. 
We return to this question in later subsections, where we dissect 
the role of task diversity, grounding, and hypernetwork conditioning. 

\subsection{Can we directly build policies from language, or do we need grounding?}

The results in Figure~\ref{fig:main-results} reveal a mixed picture. 
Direct TeNet already provides a substantial improvement over DT across all benchmarks, 
confirming that natural language is an effective source of task signals. 
However, its relative performance compared to Prompt-DT depends critically on the setting. 
On \textbf{meta-learning benchmarks} (HalfCheetah-Vel, Ant-Dir, ML1 Pick-Place), 
Direct TeNet falls behind Prompt-DT, suggesting that text encodings, while informative, 
do not generalize to unseen tasks as effectively as trajectory prompts. 
In contrast, on \textbf{multi-task benchmarks} (MT10, MT50), Direct TeNet consistently 
outperforms Prompt-DT. These results indicate that \emph{direct language-to-policy 
instantiation is viable} and scales well in diverse multi-task regimes, but that 
\emph{additional grounding is required for robust generalization} in meta-learning 
settings where the agent must extrapolate to unseen tasks.

\subsection{How should we ground language in behavior?}

The results in Figure~\ref{fig:main-results} show that grounded TeNet, regardless of the chosen 
strategy, consistently outperforms Direct TeNet on the meta-learning benchmarks 
(HalfCheetah-Vel, Ant-Dir, ML1 Pick-Place). This confirms that additional grounding is 
necessary for robust generalization to unseen tasks. 

Among the grounding methods, \textbf{contrastive alignment} generally performs better than 
direct alignment (MSE). The reason is that MSE enforces absolute closeness 
between paired text and trajectory embeddings, but provides no mechanism to separate embeddings 
from different tasks. As a result, embeddings from similar descriptions may collapse, limiting 
discriminability. In contrast, contrastive objectives simultaneously \emph{pull together} 
matching text–trajectory pairs and \emph{push apart} non-matching pairs, yielding a representation 
space that is both semantically aligned and better separated across tasks. This improved structure 
in the shared embedding space translates into stronger policy generalization.

\subsection{Why does Prompt-DT struggle in MT10 and MT50?}

The Meta-World multi-task benchmarks (MT10 and MT50) contain tasks that are far more 
distinct than those in Mujoco (e.g., pick-place versus drawer-open, compared to velocity 
or direction variations). This task diversity poses a major challenge for Prompt-DT. 
Furthermore, as the number of tasks increases, the success rate of Prompt-DT drops 
(from 0.73 on MT10 to 0.61 on MT50; see Figure~\ref{fig:main-results}). 
To better understand this gap, we conduct two follow-up experiments. 

First, we ask whether the failure is simply due to \emph{insufficient model capacity}. 
If trajectory prompts are expressive enough, then increasing the size of Prompt-DT 
(from small to medium to large) should yield meaningful improvements. 
Table~\ref{tab:mt10-mt50-hn} shows that this is not the case: larger Prompt-DT models 
achieve only marginal gains, indicating that the issue lies deeper than model capacity. 

Second, we test whether the limitation arises from the lack of \emph{task-specific 
parameterization}. In this variant, Prompt-DT-HN serves as a trajectory-conditioned hypernetwork baseline, where the prompt trajectory is encoded and used to generate policy weights via a shared hypernetwork. To this end, we add a hypernetwork on top of Prompt-DT to generate 
policy parameters conditioned on task signals. Table~\ref{tab:mt10-mt50-hn} indicates 
that this modification yields a substantial boost in success rates on both MT10 and MT50. 
The comparison demonstrates that explicitly generating task-specific parameters is 
crucial when dealing with distinct multi-task benchmarks. TeNet naturally benefits 
from this principle while also being language-enabled, removing the reliance on 
demonstration prompts.

\begin{table}[t]
    \centering
    \caption{Success rate on MT10 and MT50, along with controller size and
    control frequency. Prompt-DT-S is the default configuration.}
    \label{tab:mt10-mt50-hn}
    \small
    \begin{adjustbox}{max width=\columnwidth}
    \begin{tabular}{lccccc}
        \toprule
        \multirow{2}{*}{\textbf{Model}} &
        \multicolumn{2}{c}{\textbf{Success Rate}} &
        \multirow{2}{*}{\makecell{\textbf{Ctrl}\\\textbf{Size}}} &
        \multirow{2}{*}{\makecell{\textbf{Ctrl}\\\textbf{Freq.}}} \\
        \cmidrule(lr){2-3}
         & MT10 & MT50 & & \\
        \midrule
        Prompt-DT-S  & 0.73 & 0.61 & 1M   & 557 Hz \\
        Prompt-DT-M  & 0.79 & 0.65 & 6M   & 331 Hz \\
        Prompt-DT-L  & 0.74 & 0.58 & 39M  & 190 Hz \\
        Prompt-DT-HN & 0.99 & 0.97 & 5M   & 462 Hz \\
        TeNet        & \textbf{0.99} & \textbf{0.98} & \textbf{40K} & \textbf{9300 Hz} \\
        \bottomrule
    \end{tabular}
    \end{adjustbox}
\end{table}

\subsection{How fast are TeNet policies?}

Beyond task success, deployability depends critically on the efficiency of the policy: 
controllers must be compact enough to fit on resource-constrained robots, and fast 
enough to support high-frequency control loops. Table~\ref{tab:mt10-mt50-hn} reports 
both the number of parameters (controller size) and the control frequency that the  
method can sustain. 

The results highlight a stark contrast. Prompt-DT variants range from 1M to 39M 
parameters, with control frequencies between 190~Hz and 600~Hz. Adding a hypernetwork 
further increases model size to 5M parameters, while improving task success, but the 
resulting policies remain limited to the sub-kHz regime. In contrast, TeNet policies 
contain only \textbf{40K parameters} and sustain control rates of over \textbf{9~kHz}, 
more than an order of magnitude faster than all Prompt-DT baselines. 

% These results demonstrate that TeNet not only matches or exceeds success rates but also 
% provides \emph{lightweight and high-frequency controllers}, making it well-suited for 
% deployment on real robots where hardware constraints and responsiveness are critical. 

\subsection{Does scaling the number of training tasks improve TeNet’s generalization?}
\label{sec:task_scaling}

\begin{figure*}[t]
    \centering

    % --- Left: ML1 scaling plot ---
    \begin{minipage}[t]{0.32\textwidth}
        \vspace{0pt}
        \centering
        \includegraphics[width=0.9\linewidth]{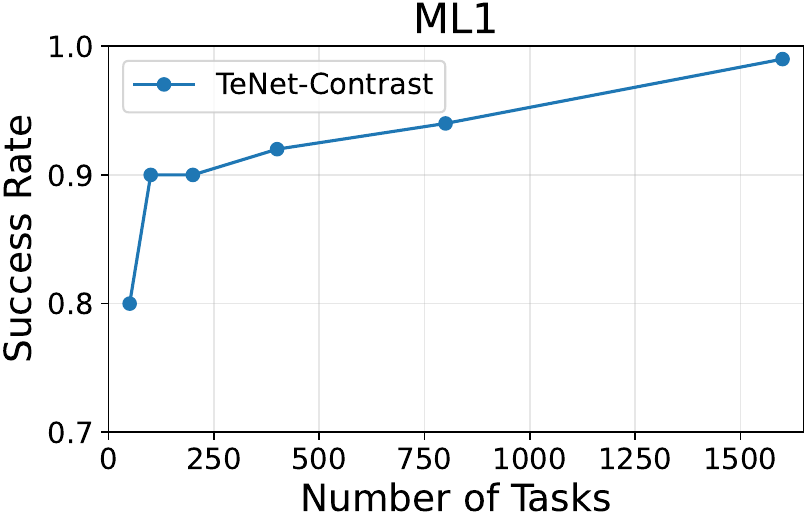}
        \captionof{figure}{TeNet-Contrast performance on ML1 Pick-Place with varying numbers of tasks.}
        \label{fig:ml1-manytask}
    \end{minipage}
    \hfill
    % --- Middle: Cheetah velocity plot ---
    \begin{minipage}[t]{0.32\textwidth}
        \vspace{0pt}
        \centering
        \includegraphics[width=0.9\linewidth]{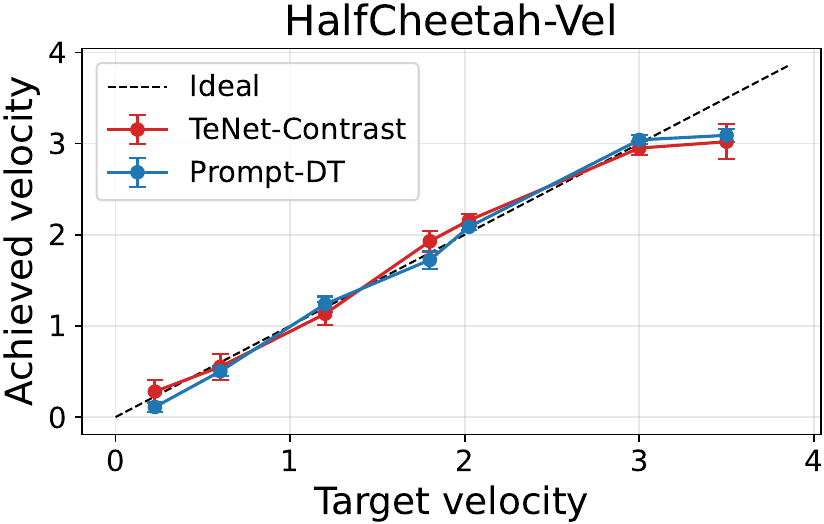}
        \captionof{figure}{Achieved forward velocity vs.\ instructed target velocity in HalfCheetah-Vel (mean over 50 rollouts).}
        \label{fig:cheetah-vel-alignment}
    \end{minipage}
    \hfill
    % --- Right: Encoder paraphrasing table ---
    \begin{minipage}[t]{0.25\textwidth}
        \vspace{8pt}
        \centering
        \captionof{table}{Success rates on MT10 when training on canonical descriptions and evaluating on increasingly complex paraphrases.}
        \label{tab:ablation-encoder}
        \vspace{0.4em}
        \begin{tabular}{lccc}
        \toprule
        \textbf{Encoder} & \textbf{L0} & \textbf{L1} & \textbf{L2} \\
        \midrule
        \textbf{LLaMA} & 0.99 & 0.95 & 0.89 \\
        \textbf{BERT}  & 0.99 & 0.89 & 0.82 \\
        \bottomrule
        \end{tabular}
    \end{minipage}

\end{figure*}

In Section~\ref{sec:main-results} we noted that TeNet-Contrast slightly underperforms 
Prompt-DT on ML1 Pick-Place. To investigate further, we study how scaling the number 
of training tasks affects generalization. Specifically, we vary the number of ML1 
tasks available during training (50, 100, 200, 400, 800, 1600), while always holding 
out 10\% of tasks for testing. The results are shown in Figure~\ref{fig:ml1-manytask}. 

Performance improves steadily from a success rate of 0.80 with 50 tasks to 0.99 with 
1600 tasks. This indicates that scaling the diversity of training tasks substantially 
enhances TeNet’s ability to generalize. One possible factor is that as the number of 
training tasks grows, the domain gap between train and test tasks decreases, making 
generalization easier. In any case, reaching a success rate of \textbf{99\%} with 
1600 training tasks shows that TeNet can fully solve ML1 Pick-Place when provided 
with sufficient data. These results highlight both the promise and the data demands 
of language-enabled hypernetworks: like foundation models in other domains, TeNet 
benefits strongly from scale, even if it is data hungry. 

% \begin{figure}[t]
% \centering
%     \includegraphics[width=0.8\columnwidth]{ml1_manytask.pdf}
%     \caption{TeNet-Contrast performance on ML1 Pick-Place with varying numbers of tasks.}
%     \label{fig:ml1-manytask}
% \end{figure}

\subsection{How Robust Is TeNet to Paraphrased Task Descriptions?}
\label{app:ablation-encoder}

Since language is used only once to instantiate a policy, a key question is how robust TeNet is to variations in how a task is described. In practice, semantically identical instructions may differ substantially in wording, syntax, or length. We therefore evaluate TeNet’s sensitivity to paraphrasing by training on canonical task descriptions and testing on increasingly complex paraphrases. We conduct this study on MT10, comparing two text encoders—LLaMA~\cite{touvron2023llama} and BERT~\cite{devlin2019bert}. Models are trained using 10 canonical (Level~0) descriptions per task and evaluated on unseen paraphrases of growing linguistic complexity (Level~1 and Level~2), which differ syntactically but describe the same underlying task.

% \begin{table}[h!]
% \centering
% \caption{Success rates on MT10 when training on canonical descriptions and evaluating on increasingly complex paraphrases.}
% \label{tab:ablation-encoder}
% \begin{tabular}{lccc}
% \toprule
% \textbf{Encoder} & \textbf{Level 0} & \textbf{Level 1} & \textbf{Level 2} \\
% \midrule
% \textbf{LLaMA} & 0.99 & 0.95 & 0.89 \\
% \textbf{BERT}  & 0.99 & 0.89 & 0.82 \\
% \bottomrule
% \end{tabular}
% \end{table}
Both encoders achieve identical performance on canonical descriptions, indicating that TeNet can reliably instantiate policies from simple instructions regardless of the encoder choice. However, as paraphrasing complexity increases, performance degrades for both models, with a substantially larger drop observed for BERT. This gap suggests that richer language models produce more stable and semantically consistent embeddings under linguistic variation, leading to more reliable policy instantiation from natural language.

% \begin{figure}[t]
%     \centering
%     \includegraphics[width=0.85\linewidth]{cheetah_vel_target_vs_achieved}
%     \caption{
%         Achieved forward velocity as a function of instructed target velocity in HalfCheetah-Vel.
%         Points show mean achieved speed over $50$ rollouts for each instruction.
%     }
%     \label{fig:cheetah-vel-alignment}
% \end{figure}

\subsection{Qualitative Evaluation: Velocity Following in HalfCheetah-Vel}
\label{app:cheetah-vel-alignment}

HalfCheetah-Vel evaluates velocity tracking by rewarding policies for matching a target forward speed, $r = -\left| v_{\text{current}} - v_{\text{target}} \right|$, 
so episodic return directly reflects tracking accuracy. Target velocities are defined on a grid from $0.075$ to $3.0\,\text{m/s}$. Models are trained on a subset of this grid and evaluated on unseen target velocities
$\{0.225,\ 0.6,\ 1.2,\ 1.8,\ 2.025\}\,\text{m/s}$, as well as an out-of-distribution instruction at $3.5\,\text{m/s}$. Policies are instantiated from commands of the form
\emph{``Move forward with target velocity $X$ m/s.''}
and evaluated over 50 rollouts. Achieved velocity is computed as the average forward speed over the final 20 steps. 

Figure~\ref{fig:cheetah-vel-alignment} compares achieved versus instructed velocity for TeNet-Contrast and Prompt-DT. TeNet-Contrast closely tracks commanded speeds across all unseen test velocities, indicating smooth generalization over the continuous velocity range. For the extrapolated $3.5\,\text{m/s}$ instruction, both methods saturate near $\sim 3\,\text{m/s}$, reflecting the physical limits of the HalfCheetah dynamics rather than a failure of instruction following.

\subsection{Summary of Empirical Insights}

Our results show that direct text-to-policy instantiation is viable, but that grounding language in behavior is essential for robust generalization. Across meta-learning benchmarks, grounded TeNet variants consistently outperform Direct TeNet, with contrastive alignment providing stronger task discrimination than direct (MSE) alignment.

We further find that task-specific parameterization is critical in diverse multi-task settings. Prompt-DT degrades sharply on MT10 and MT50, and increasing model capacity alone does not resolve this issue. In contrast, explicitly generating task-conditioned parameters—most effectively via language-conditioned hypernetworks—enables TeNet to scale to large and heterogeneous task sets.

Despite this added flexibility, TeNet remains highly efficient. Instantiated policies contain only $\sim$40K parameters and sustain control rates above 9~kHz, exceeding Prompt-DT baselines by more than an order of magnitude. TeNet also exhibits robustness to linguistic variation: performance degrades gracefully under increasingly complex paraphrases, with larger language models such as LLaMA producing more stable policy instantiations than BERT. Qualitative evaluation on HalfCheetah-Vel further confirms that text-instantiated policies accurately follow commanded velocities and generalize smoothly across unseen targets.

Additional ablation studies—analyzing the effect of the text--text contrastive term, grounded-flow, fine-tuning, and multiple task descriptions—are provided in the supplementary materials.

Overall, these findings indicate that compact, language-enabled hypernetworks can close much of the gap between lightweight sequence models and large language-conditioned systems within state-based, offline imitation settings. Extending TeNet to real-world robotics will require addressing noisy demonstrations, multimodal (vision--language) grounding, and reinforcement fine-tuning, which we leave as directions for future work.

\section{Conclusion}

We introduced TeNet, a text-to-network framework for instantiating compact, task-specific 
policies directly from natural language. By combining LLM-based text embeddings, 
trajectory grounding, and hypernetwork-based parameter generation, TeNet produces 
lightweight controllers that generalize across tasks without requiring demonstrations 
at inference time. Experiments on Mujoco and Meta-World benchmarks show that TeNet 
outperforms Prompt-DT in multi-task settings, achieves competitive performance in 
meta-learning, and supports control frequencies above 9~kHz. Together, these results 
highlight language-enabled hypernetworks as a promising direction for scalable, 
efficient, and deployable robot learning.

\bibliographystyle{named}
\bibliography{ijcai26}

\end{document}